\documentclass[11pt,a4paper]{article}
\usepackage{times,latexsym}
\usepackage{url}
\usepackage[T1]{fontenc}
\usepackage{amsmath,amssymb}
\usepackage{amsthm}
\usepackage{booktabs}
\usepackage{multirow}
\usepackage{multicol}
\usepackage{graphicx}
\usepackage[table]{xcolor}

\usepackage{caption}
\usepackage{hyperref}
\usepackage[noabbrev]{cleveref}
\usepackage{subcaption}

\usepackage[acceptedWithA]{tacl2021v1}

\usepackage{xspace,mfirstuc,tabulary}

\newif\iftaclinstructions
\taclinstructionsfalse 
\iftaclinstructions
\renewcommand{\confidential}{}
\renewcommand{\anonsubtext}{(No author info supplied here, for consistency with
TACL-submission anonymization requirements)}
\newcommand{\instr}
\fi

\iftaclpubformat 

\else

\fi

\newtheorem{definition}{Definition}

\DeclareMathOperator{\E}{\mathbb{E}}
\DeclareMathOperator{\trace}{Tr}

\DeclareMathOperator*{\softmax}{softmax}

\newcommand{\jachess}[0]{\textsc{JacHess}}

\title{From Robustness to Improved Generalization and Calibration in Pre-trained Language Models}

\author{Josip Juki{\'{c}} \quad Jan {\v{S}}najder\\
TakeLab\\
Faculty of Electrical Engineering and Computing\\
University of Zagreb, Croatia\\
\tt \{josip.jukic, jan.snajder\}@fer.hr
} 

\date{}

\begin{document}

\maketitle


\begin{abstract}

Enhancing generalization and uncertainty quantification in pre-trained language models (PLMs) is crucial for their effectiveness and reliability. Building on machine learning research that established the importance of robustness for improving generalization, we investigate the role of representation smoothness, achieved via Jacobian and Hessian regularization, in enhancing PLM performance. Although such regularization methods have proven effective in computer vision, their application in natural language processing (NLP), where PLM inputs are derived from a discrete domain, poses unique challenges. We introduce a novel two-phase regularization approach, \jachess{}, which minimizes the norms of the Jacobian and Hessian matrices within PLM intermediate representations relative to their inputs. Our evaluation using the GLUE benchmark demonstrates that \jachess{} significantly improves in-domain generalization and calibration in PLMs, outperforming unregularized fine-tuning and other similar regularization methods.

\end{abstract}

\section{Introduction}


Effective generalization, broadly understood as the capability to transfer learned representations, knowledge, and strategies from familiar contexts to new ones, stands out as a key goal for models in natural language processing (NLP) and extends to the broader domain of machine learning \cite{hupkes-etal-2023-taxonomy}. The standard approach to achieve generalization in machine learning hinges on minimizing the empirical risk, which serves as a surrogate for the true risk. Such an approach gives rise to the notion of the \textit{generalization gap}, the discrepancy between the empirical risk and the true risk, which can be gauged through model properties. Notably, \textit{robustness}, defined as the amount that the loss can vary with respect to changes in the \textit{inputs}, offers an effective lens for understanding generalization \cite{vapnik-1998-statistical, deng-etal-2021-toward, zhang-etal-2021-understanding}. Recently, \citet{kawaguchi-etal-2022-robustness} provided tighter generalization bounds for robustness, a challenge that has persisted since robustness was first proposed as an approach for analyzing learning algorithms \cite{xu-mannor-2012-robustness}. The drawback of their approach is that the bounds are data-dependent, meaning that the exact values of the bounds remain undetermined until the training data is specified, highlighting an area ripe for further exploration in the field.


Robustness in neural networks is intrinsically linked to the concept of \textit{representation smoothness}. When representation space is smooth, it ensures that outputs vary minimally in response to small input changes, thereby maintaining a stable loss by controlling the network's geometric complexity \cite{dherin-etal-2022-neural}. Promoting smoothness in representations not only aids in better generalization but also supports more reliable uncertainty quantification in the model's predictions \cite{rosca-etal-2020-case}, a critical area where neural networks often fall short. Typically, this shortfall is seen as overconfidence in the model's predictions, resulting in poor calibration --- the alignment between the predicted probabilities and the true probabilities of outcomes \cite{guo-etal-2017-calibration}. Therefore, promoting smooth representations emerges as a promising strategy for boosting generalization through robustness and stabilizing uncertainty quantification in neural networks.


While traditional regularization techniques such as weight decay and dropout are widely recognized as generalization enhancers in neural networks \cite{loshchilov-and-hutter-2018-decoupled, srivastava-etal-2014-dropout}, they do not fully address the nuances of robustness in the representation space against input changes. In contrast, specialized strategies aimed at boosting robustness provide a more focused solution. These methods manipulate the input-output \textbf{Jacobian} and \textbf{Hessian} matrices, which involve first-order and second-order partial derivatives, respectively. Minimizing the norms of these matrices increases the network's representation smoothness, enhancing its robustness. This approach has proven highly successful in the field of computer vision, as evidenced by several studies \cite{czarnecki-etal-2017-sobolev, varga-etal-2018-gradient, sokolic-etal-2017-robust, mustafa-etal-2020-input}. From the viewpoint of function space theory, a neural network can be viewed as a function where the principle of \textit{Lipschitz continuity} serves as a reflection of its smoothness against input variations, indicating the maximum rate of change in the function's output relative to a change in the input \cite{khromov-and-sidak-2024-some}. Reducing the network's Lipschitz constant thus reduces its sensitivity to input perturbation. While computing the exact Lipschitz constant is an NP-hard problem \cite{virmaux-etal-2018-lipschitz}, leveraging Jacobian and Hessian norms as proxies can provide a practical approach to emulate these effects, further advancing the robustness of neural networks.


Directly addressing robustness via representation smoothness presents a refined strategy over traditional regularization methods, emphasizing the importance of directly addressing robustness for improved generalization. While these methods have been shown to work well in computer vision, none of them have been used in NLP, revealing a significant research gap. Specifically, there is untapped potential to apply representation-based regularization methods to enhance generalization in pre-trained language models (PLMs) \cite{liu-etal-2023-pac}, which currently play a pivotal role in advancing NLP. However, a notable challenge in this research direction, unlike in computer vision, is the discrete nature of tokens that PLMs must process. This issue echoes previous challenges encountered when methods proven effective in computer vision did not seamlessly transfer to the language processing field (e.g., the idea of generative adversarial networks that operate by applying slight continuous modifications to inputs \cite{goodfellow-etal-2020-generative}).


 In this paper, we adapt and expand upon the regularization techniques used in the representation space within computer vision, applying them to NLP. While the challenges posed by the discrete nature of data processed by PLMs are clear, we find a workaround in leveraging the \textbf{continuous embedding space} as a viable substitute. We introduce \jachess{}, a novel regularization approach that minimizes the norms of the Jacobian and Hessian matrices within PLM representations relative to their inputs. By targeting both Jacobian and Hessian norms, \jachess{} doubly enhances model robustness by reducing sensitivity to input changes and smoothing the curvature of representations. Typically, calculating these norms in high-dimensional settings is resource-intensive, but we address this through the use of estimators. Inspired by the goal of achieving smooth transitions in the features learned across intermediate layers \cite{lange-etal-2022-neural, he-and-su-2023-law}, our method applies regularization across the network's layers. Moreover, \jachess{} employs a two-phase strategy: iterative fine-tuning using labeled data followed by regularization with additional \textbf{unlabeled} data.
 In standard scenarios, applying \jachess{} to the training data already improves generalization. However, applying regularization to a separate set of unlabeled data --- which is typically much more accessible than extra labeled data --- significantly boosts the model's generalization capabilities.
 

 Given their growing importance, our evaluation concentrates on decoder-based models, including the OPT family and LLaMA-2, where we fine-tune the PLMs and evaluate them on the GLUE benchmark \cite{wang-etal-2018-glue}. We validate \jachess{} by examining PLM robustness against the perturbations in embedding space and its impact on handling corrupted token inputs. Subsequently, we assess the effectiveness of \jachess{} in improving models' predictive accuracy and uncertainty quantification. Our findings show that \jachess{} outperforms standard unregularized fine-tuning and other Jacobian- and Hessian-based regularization methods, markedly improving generalization and uncertainty quantification through smoothing out representations.


In summary, our work presents two significant contributions:
\begin{enumerate}
    \item We conduct an empirical assessment of enforcing representation smoothness on PLMs via Jacobian- and Hessian-based regularization.
    \item We introduce \jachess{}, a novel regularization approach that not only improves model generalization beyond the capabilities of standard fine-tuning and other regularization methods in the representation space but also improves uncertainty quantification.
\end{enumerate}
The implications of this work set a new standard for enhancing the robustness, generalization, and reliability of PLMs in the ever-evolving landscape of NLP.

\section{Related Work}

\paragraph{Generalization enhancers.}
From the standpoint of machine learning theory, regularization is commonly employed during training to boost generalization \cite{liu-etal-2023-pac}. Although increasing the amount of labeled data can straightforwardly improve generalization, the challenge lies in optimizing the use of existing data or in capitalizing on the plentiful supply of unlabeled data. Within the domain of PLMs, various strategies aim to address this challenge. Notably, \citet{gururangan-etal-2020-dont} proposed task-adaptive pre-training, which involves additionally training the model on the task-specific unlabeled set via the language modeling task. In a different approach, \citet{yu-etal-2021-fine} leveraged unlabeled data for generalization enhancement through the mechanism of weak supervision. Moreover, beyond the utilization of unlabeled data, the field has seen advancements in generalization through data augmentation methods \cite{okimura-etal-2022-impact, zhou-etal-2021-flipda, wu-etal-2022-text}.

\paragraph{Representation smoothness.}
Examining the impact of input modifications on a function's output is essential for enhancing the generalization capabilities of neural networks. \citet{rosca-etal-2020-case} defined this concept as \textit{model smoothness}, which can be referred to as representation smoothness in the context of deep neural networks. They advocate for implementing smoothness constraints with respect to inputs to improve cross-domain and robust generalization and provide reliable uncertainty quantification. While numerous theoretical approaches predominantly focus on Lipschitz-bounds to assess model sensitivity \cite{bartlett-etal-2017-spectrally, bubeck-etal-2021-a, wang-etal-2023-direct}, it has been demonstrated that Lipschitz continuity can be effectively leveraged for regularization purposes \cite{gouk-etal-2021-regularisation}.

\paragraph{Robustness and generalization.}
Research into the effects of Jacobian regularization was first explored in \cite{drucker-and-lecun-1992-improving}. Since then, several iterations of this technique have been proposed \cite{sokolic-etal-2017-robust, ororbia-etal-2017-unifying}, primarily aimed at enhancing robustness against adversarial examples \cite{schmidt-etal-2018-adversarially, li-etal-2022-robust}. Extending this work, \citet{varga-etal-2018-gradient} suggested that such robustness could also benefit standard, in-domain generalization. Further developments by \citet{hoffman-etal-2019-robust} introduced an estimator for the norm of the Jacobian matrix of logits with respect to inputs in order to avoid computing the whole Jacobian matrix, which is computationally expensive. This idea enabled a more resource-efficient regularization technique by minimizing the norm in the representation space with respect to the inputs. Expanding the scope, \citet{mustafa-etal-2020-input} extended this concept to regularizing the norm of the Hessian matrix, focusing on the scalar output of binary classification scores. In our research, we push beyond these established boundaries and apply regularization in the representation spaces of PLMs, advocating for the promotion of smoothness across intermediate representations throughout the layers. We also explore applying regularization solely on separate, unlabeled data, marking a departure from traditional techniques that use the training set.
\section{Methodology}

In this section, we outline our methodology, starting with the theoretical background of \textit{Lipschitz continuity}, a valuable concept for assessing the stability and smoothness of neural networks. We then address computational challenges and propose strategies to mitigate them. Finally, we introduce our regularization technique, \jachess{}, which leverages the continuous embedding space of PLMs to employ representation-based regularization with respect to the inputs.

\subsection{Lipschitz continuity}

From the perspective of function space analysis \cite{sonoda-noboru-2017-neural}, the mapping of a neural network from an $n$-dimensional input to its $m$-dimensional output of any given layer can be represented as a vector-valued function of a vector variable $f: \mathbb{R}^n \rightarrow \mathbb{R}^m$. The Lipschitz continuity of a function is a crucial property that indicates its degree of smoothness when the input is altered. More precisely, it represents the greatest absolute rate of change in the function's output for a unit norm change in its input. In the context of machine learning, we prefer our learned predictive function to be robust; that is, it should not be excessively sensitive to input variations \cite{khromov-and-sidak-2024-some}.
\begin{definition}[$L$-Lipschitz continuity]
A function $f : \mathbb{R}^n \rightarrow \mathbb{R}^m$ is termed \textit{$L$-Lipschitz continuous} if there is a real constant $L \geq 0$ such that:
\[
\|f(\mathbf{x}) - f(\mathbf{x}')\|_2 \leq L \|\mathbf{x} - \mathbf{x}'\|_2,
\forall \mathbf{x}, \mathbf{x}' \in \mathbb{R}^n \text{.}
\]
\end{definition}

Importantly, there is a link between the Lipschitz constant and the \textbf{Jacobian} matrix, which is often exploited when estimating the smoothness of functions implemented by neural networks. Let \( \mathbf{J}_f(\mathbf{x}) : \mathbb{R}^n \rightarrow \mathbb{R}^{m \times n} \) denote the Jacobian matrix of \( f \), whose elements are:
\[
\left[\mathbf{J}_f(\mathbf{x})\right]_{i,j} = \frac{\partial}{\partial x_j}f_i(\mathbf{x}).
\]
The spectral norm\footnote{The spectral norm of a matrix is defined as $\|\mathbf{A}\|_2 = \max_{\mathbf{x} \neq \mathbf{0}} \frac{\|\mathbf{A}\mathbf{x}\|_2}{\|\mathbf{x}\|_2}$. This norm is equivalent to the maximum singular value of $\mathbf{A}$.} of $\mathbf{J}_f(\mathbf{x})$ is constrained by the function's Lipschitz constant $L$, effectively serving as a lower bound \cite{dherin-etal-2022-neural}:
\begin{equation*}
   \|\mathbf{J}_f(\mathbf{x})\|_2 \leq L, \forall \mathbf{x} \in \mathbb{R}^n \text{.}  
\label{eq:lb}
\end{equation*}
If we consider the relationship between matrix norms, specifically for any matrix $\mathbf{A}$ of rank $r$ over a field of real or complex numbers, we have:
\begin{equation*}
    \|\mathbf{A}\|_2 \leq \|\mathbf{A}\|_F \leq \sqrt{r} \|\mathbf{A}\|_2 \text{,}
\end{equation*}
where $\|\cdot\|_F$ is the Frobenius norm. This suggests that reducing the Frobenius norm can lead to a decreased spectral norm, thereby fostering smoother function behavior. This is corroborated by evidence showing that the Lipschitz constant of neural networks closely follows the lower bound defined by \cref{eq:lb} \cite{latorre-etal-2020-lipschitz, khromov-and-sidak-2024-some}.

\begin{definition}[$L$-Lipschitz gradient continuity]
If we decompose the Jacobian matrix as a set of gradients of function $f$ for each of its output dimension $j$, then the $i$-th dimension $f_j$ has a gradient that is $L$-Lipschitz continuous for some $L > 0$ if
\[
\| \nabla f(\mathbf{x}) - \nabla f(\mathbf{x}') \|_2 \leq L\|\mathbf{x} - \mathbf{x}'\|_2, \forall \mathbf{x}, \mathbf{x}' \in \mathbb{R}^n \text{.}
\]
We call such functions $L$-Lipschitz smooth.
\end{definition}

A function that is $L$-Lipschitz limits how rapidly its output can change with respect to changes in the input. By applying this constraint to the function's gradients, we establish that they cannot vary sharply and must be bound by a specific value, as defined above. Put another way, $L$-smoothness imposes an upper limit on the curvature of the function, which is inherently linked to the \textbf{Hessian} matrix $\mathbf{H}_f(\mathbf{x})$. The Hessian matrix consists of second-order derivatives and is defined for functions with scalar outputs. In the case of a vector-valued function, we can decompose the vector output into scalars where each scalar output has its corresponding Hessian matrix. Generally, the Hessian matrix is a square matrix \( \mathbf{H}_f(\mathbf{x}): \mathbb{R}^n \rightarrow \mathbb{R}^{n \times n} \), whose elements are:
\[
\left[\mathbf{H}_f(\mathbf{x})\right]_{i,j} = \frac{\partial ^2 f(\mathbf{x})}{\partial x_i \partial x_j} \text{.}
\]
L-smoothness is equivalent to the condition that the eigenvalues of the Hessian matrix are smaller than $L$, which is also bounded by the spectral norm. Again, we can minimize the Frobenius norm to promote minimizing the spectral norm. This will come in handy later when we deal with norm estimations, where the \textit{Frobenius norm is much less expensive to compute than the spectral norm}.


\subsection{Hutchinson's estimator}
Computing the norms of Jacobian and Hessian matrices is computationally demanding and often infeasible for large-scale neural networks. Hutchinson's estimator \cite{hutchinson-1989-stochastic} presents a solution by enabling the computation of these norms without requiring the full matrices, thereby achieving feasible computation times. Hutchinson’s estimator provides a means to compute the trace of a matrix without needing to calculate the full matrix, which is essential for efficient computation. It uses random projections to leverage the quick computation of the gradient of a vector-matrix product:
\begin{equation}
    \E[\mathbf{v}^\top \mathbf{A} \mathbf{v}] = \trace(\mathbf{A}) \text{,}
\label{eq:fr_trace}
\end{equation}
where $\mathbf{v} \sim \mathcal{N}(\mathbf{0},\mathbf{I})$ and $\mathbf{A}$ is an arbitrary square matrix. Given the connection between the trace and the Frobenius norm noted in \cref{eq:fr_trace}, \citet{varga-etal-2018-gradient} and \citet{hoffman-etal-2019-robust} used random projections to estimate the Frobenius norm of the Jacobian matrix:\footnote{The Jacobian matrix is considered an operator, denoted $\mathbf{J}(\mathbf{x})$, which acts on the input $\mathbf{x}$. Here, we just use $\mathbf{J}$ as a shorthand for this operator.}
\begin{equation*}
    \E[\| \mathbf{v}^\top \mathbf{J} \|^2_2] = \E[\mathbf{v}^\top \mathbf{J} \mathbf{J}^\top \mathbf{v}] = \trace(\mathbf{J} \mathbf{J}^\top) = \|\mathbf{J}\|^2_F \text{.}
\end{equation*}
The estimate for the Jacobian norm is given by:
\begin{equation}
    \|\mathbf{J}\|^2_F \approx \frac{1}{p} \sum_{i=1}^p \left\| \frac{\partial(\mathbf{v}^i \mathbf{z})}{\partial \mathbf{x}} \right\|^2 \text{,}
\label{eq:jac}
\end{equation}
where $\mathbf{x}$ is the input, $\mathbf{z} = \mathbf{f}(\mathbf{x})$ is the output of a particular layer, $\mathbf{v}$ is a normalized unit vector from a normal distribution, and $p$ is the number of projections.
Estimating the norm of the Hessian matrix for the $j$-th output dimension is reduced to:
\begin{equation}
    \|\mathbf{H}_j\|^2_F \approx \frac{1}{p} \sum_{i=1}^p  \left\| \frac{\partial(\mathbf{v}^i \frac{\partial \mathbf{z}_j}{\partial \mathbf{x}})}{\partial \mathbf{x}}  \right\|^2 \text{,}
\end{equation}
where $\mathbf{z} = \mathbf{f}(\mathbf{x})$ and $\mathbf{v}$ is a normalized unit vector from a normal distribution.

\subsection{\jachess}

Our regularization method, \jachess{}, is designed to minimize the norms of the \textbf{Jac}obian and \textbf{Hess}ian matrices to promote robustness through Lipschitz continuity and smooth representation. \jachess{} adopts Hutchinson's estimator to compute these norms effectively, maintaining computational efficiency. To circumvent the problem of discrete input space of PLMs, we turn to the \textbf{continuous embedding space} by embedding the tokens and then using the embedding as the inputs. \jachess{} focuses on \textbf{intermediate representations} across network layers beyond the traditional scope of the logits (i.e., the penultimate layer's outputs). This approach considers the norm estimates of the Jacobian and Hessian matrices of a layer's representation in relation to the inputs. Additionally, it critically considers the high-dimensional nature of intermediate representations. Where other representation-based regularization methods limit their scope to low-dimensional outputs confined to the number of classes --- requiring just a few Hessian norm estimates --- our technique utilizes the high-dimensional space of intermediate layers. Recognizing that computing Hessian norms across all dimensions is computationally prohibitive, we integrate a dimension subsampling strategy. By selecting a random subset of representation dimensions for each batch, we reconcile the need for comprehensive regularization with the limits of computational feasibility. For a network with $K$ layers and $D^{(k)}$ as the random subset of indices for output dimensions for the $k$-th layer, we define the regularization term $\Omega_{\jachess{}}$ as:
\begin{equation}
 \sum_{k=1}^K \left( \lambda_1^{(k)} \|\mathbf{J}^{(k)}\|_F + \lambda_2^{(k)} \sum_{d \in  D^{(k)}} \|\mathbf{H}^{(k)}_{d}\|_F \right) \text{,}
\end{equation}
where $\lambda_i^{(k)}$ are regularization factors.
In addition to applying our method to the training set inputs (\jachess{}$_\text{train}$) simply by adding $\Omega_{\jachess{}}$ to the original loss function, we also explore a variant using a separate \textbf{unlabeled} dataset (\jachess{}$_\text{val}$). In the latter case, we minimize the regularization term on its own.

\paragraph{Regularization factors.}
In the process of selecting optimal regularization factors $\lambda_i^{(k)}$, we have conducted an extensive empirical investigation. The preservation of the base large language model's (PLM's) inherent smoothness emerged as a critical determinant for achieving the best fine-tuning outcomes. Initially, we compute the norms of the Jacobian for each layer of the PLM. Let $\mathbf{j} = \left[\|\mathbf{J}^{(1)}\|_F, \|\mathbf{J}^{(2)}\|_F, \dots , \|\mathbf{J}^{(K)}\|_F\right]$ represent the vector of these per-layer Jacobian norms estimated as in \cref{eq:jac}, where $\mathbf{J}^{(k)}$ is the Jacobian matrix of the $k$-th layer and $K$ is the number of layers. The corresponding vector of regularization factors $\boldsymbol{\lambda}$ is then determined by the equation:
\begin{equation}
\boldsymbol{\lambda} = \xi \softmax(-\mathbf{j}),
\end{equation}
where $\xi$ serves as a regularization constant aligned with the learning rate, ensuring that the regularization strength is scaled appropriately in relation to the learning process. This approach assigns regularization factors that are inversely related to the Jacobian norms --- essentially, smaller factors for less smooth layers --- aiming to retain the smoothness distribution present in the pre-trained model's layers and prevent the dilution of previously acquired knowledge. This specific allocation strategy for $\boldsymbol{\lambda}$ is scrutinized in an ablation study detailed in \Cref{sec:analysis}, confirming its efficacy. It's important to note that we apply identical regularization factors to both the Jacobian and Hessian terms, thereby setting $\lambda_1^{(k)} = \lambda_2^{(k)}$, which emerged as the best strategy.
\section{Experimental Setup}

This section outlines the experimental framework utilized in our study.

\subsection{Model Selection}
For our investigation, we opted to evaluate a variety of decoder-based language models due to their increasing prominence. We explored a series of decoder-based OPT models across three distinct sizes: 125 m, 1.3 b, and 6.7 b parameters. We also included the LLaMA-2 variant with 7 b parameters. As a baseline comparison, we included BERT, an encoder-based model. 

\subsection{Datasets}
Our analysis is conducted using the GLUE benchmark, following a standard practice of work on in-distribution generalization. We include four binary classification tasks for single sequences (CoLA, SST-2, and RTE), three binary classification tasks for sequence pairs (MRPC, QQP, QNLI), one multi-class classification task for sequence pairs (MNLI), and one regression task (STS-B). This selection ensures a comprehensive evaluation across a variety of linguistic challenges.

\subsection{Baselines}
In our analysis, we incorporated two regularization techniques based on the first and second order of gradient regularization. Specifically, we use Jacobian regularization \cite{hoffman-etal-2019-robust}, aimed at minimizing the gradient of the model's logits (the penultimate layer outputs) with respect to its inputs. Concurrently, we utilized Cross-Hölder regularization \cite{mustafa-etal-2020-input}, which combines the gradient norm of the logits with respect to inputs and the norm of the associated Hessian matrix (the second-order derivatives).

\subsection{Fine-tuning}

We fine-tune the language model on each task, capping the training set at $10,000$ instances. This cap is implemented to enhance computational efficiency, as performance typically plateaus on the GLUE benchmark when the number of training instances exceeds $10,000$. With encoder models, the [CLS] token's representation is employed, while for decoders, we rely on the representation of the layer's last token. In scenarios involving sequence pairs, we merge the encoder model representations with a [SEP] token in between, and for decoders, a [EOS] token is added in between the representations. We stack a linear head on top of the final layer, and we proceed to fine-tune the entire model. An exception is made for the larger models in our study, specifically OPT-6.7b and LLaMa-2-7b, which are fine-tuned using the LoRA technique to enhance parameter efficiency.\footnote{We indicate this in the experiments with a star at the end: OPT-6.7b* and LLaMa-2-7b*} This strategy helps circumvent memory constraints inherent in commodity GPU usage.

We run each experiment five times using five different random seeds and report the averages. We use $2\cdot10^{-5}$ as the learning rate for standard fine-tuning and $1\cdot10^{-4}$ for tuning with LoRA. We set the regularization constant $\xi$ to be equal to the learning rate.
\section{Enhancing Generalization through Robustness}

To deepen our understanding of how optimizing robustness in embedding space can enhance generalization, we conduct a series of experiments focusing on two key aspects: embedding perturbation and token corruption. Through these experiments, we explore the resilience of models to such disruptions, intending to establish a relationship between token-level robustness and embedding space robustness. This will demonstrate their combined effect on enhancing the generalization capabilities of the model.

\subsection{Embedding Perturbation}

\begin{figure*}[t!]
\centering
\begin{subfigure}{.32\linewidth}
  \centering
  \includegraphics[width=\linewidth]{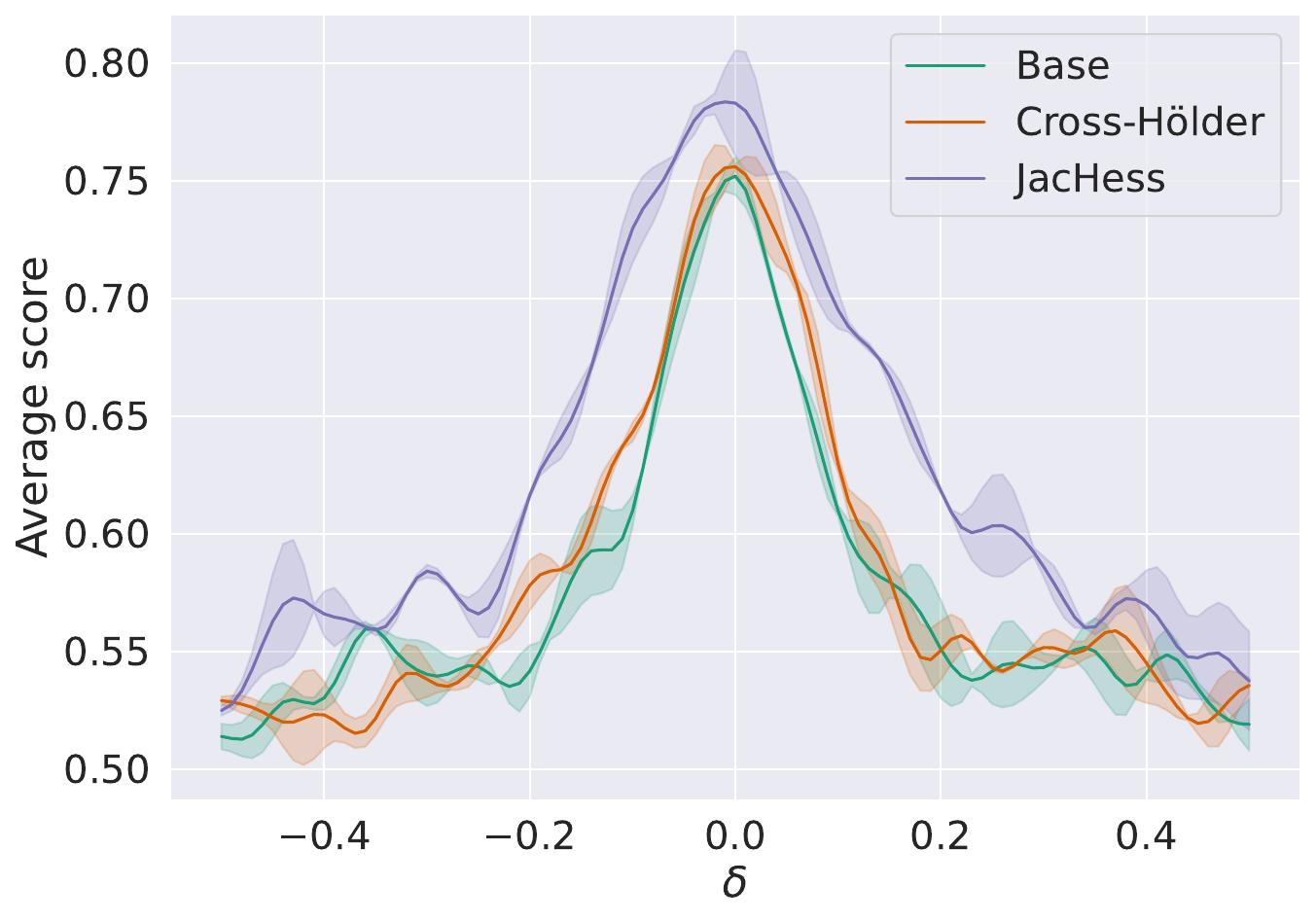}
  \caption{BERT}
\end{subfigure}
\begin{subfigure}{.32\linewidth}
  \centering
  \includegraphics[width=\linewidth]{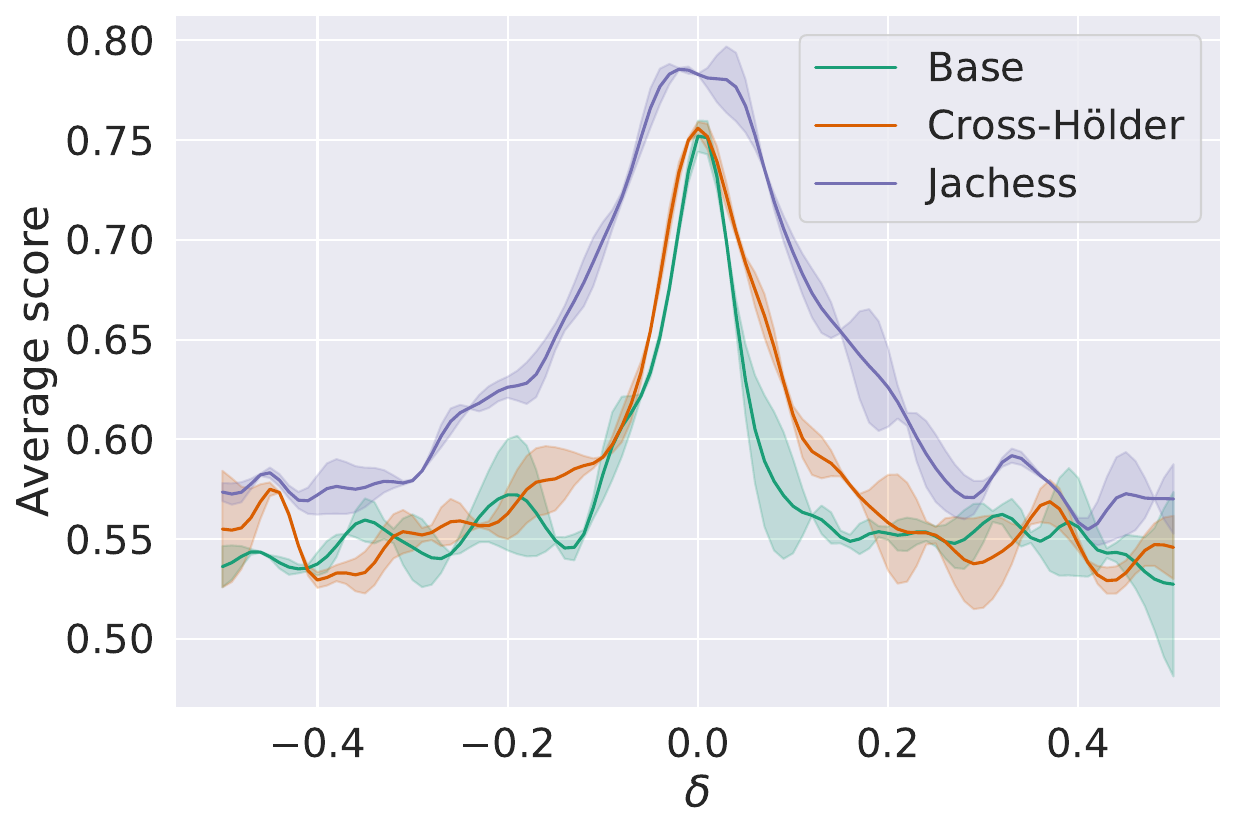}
  \caption{OPT-125m}
\end{subfigure}
\begin{subfigure}{.32\linewidth}
  \centering
  \includegraphics[width=\linewidth]{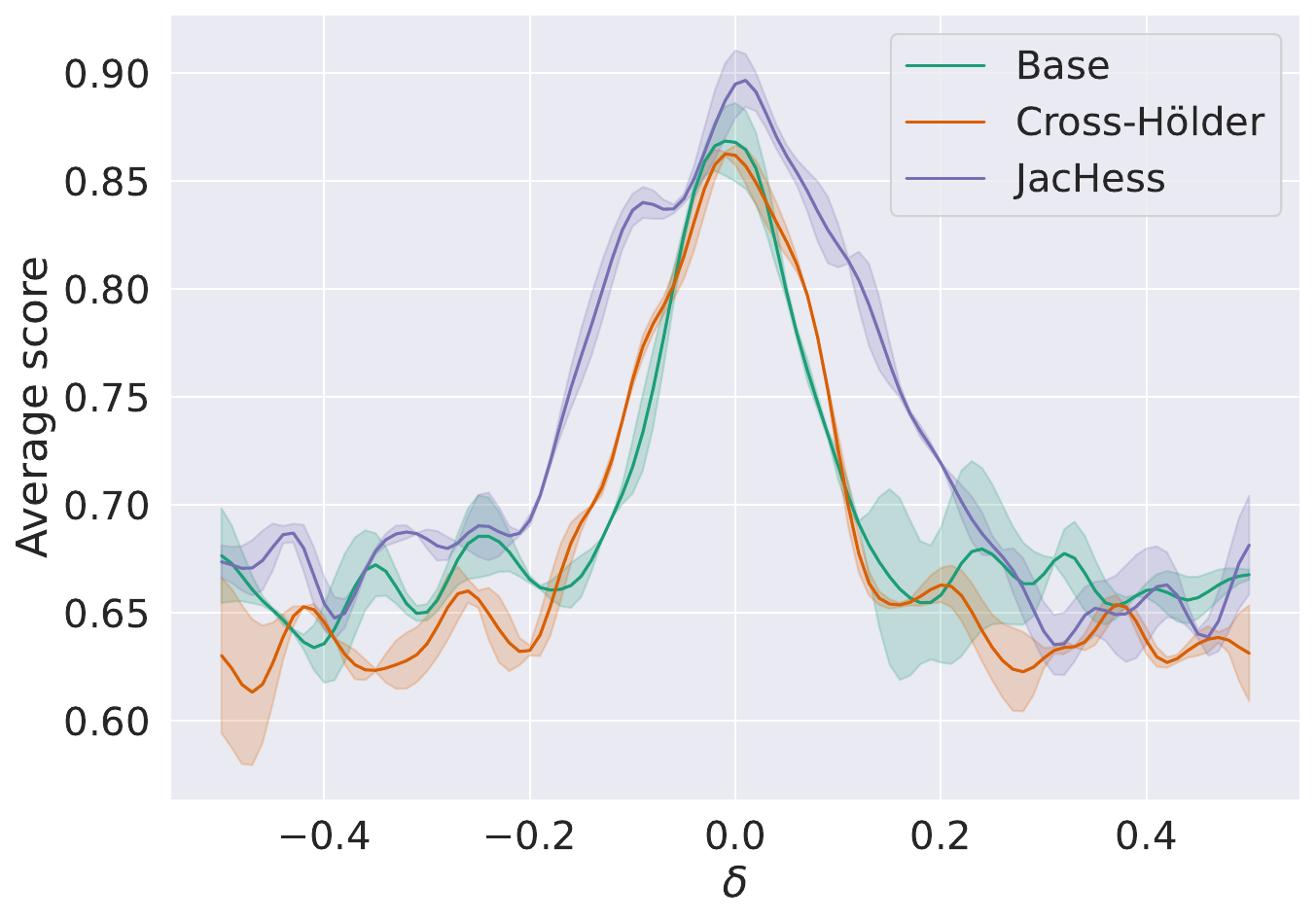}
  \caption{LLaMA}
\end{subfigure}
\caption{Average generalization score with embedding perturbation. We perturb the instances in the embedding space with factor $\delta$ for controlling the degree of perturbation. The average is reported across datasets. To avoid clutter, we plot the results for the base model without regularization, the Cross-Hölder method, which generally outperforms the Jacobian approach in this aspect, and our method, \jachess{}. Due to space constraints, we show the results for BERT, OPT-125, and LLaMA-2-7b.}
\label{fig:pert}
\end{figure*}

Our initial set of experiments targets the robustness of models to continuous perturbations in the embedding space. We introduce these perturbations by adding noise to the embeddings, specifically by augmenting the token embeddings $\mathbf{e}$ with a noise vector scaled by a factor $\delta$. The perturbed embedding $\mathbf{e'}$ is defined as:
\begin{equation}
\mathbf{e}' = \mathbf{e} + \delta \mathbf{v}, \ \mathbf{v} \sim \mathcal{N}(0,1),
\end{equation}
where $\boldsymbol{\epsilon}$ is a normally distributed random vector. The degree of perturbation is controlled by $\delta$, providing a quantitative measure of robustness in this space. Results from these perturbation tests are presented in \Cref{fig:pert}. \jachess{} maintains better generalization performance under embedding perturbation, resulting in broader generalization curves.

Despite utilizing Cross-H\"{o}lder to regularize with respect to the embedded inputs, we observe no significant enhancement in robustness against perturbed embeddings compared to the base model fine-tuned without regularization. We hypothesize that the superior robustness of \jachess{} to embedding perturbations stems primarily from \textit{applying regularization throughout the network's layers}.

\subsection{Token Corruption}

Building on the insights from embedding perturbations, we next examine how models fine-tuned with various regularization strategies endure discrete token corruption. This step is crucial for understanding the translation of robustness from the embedding space to the more granular level of tokens. We introduce corruption at varying rates ($10$\%, $15$\%, and $20$\%) to the tokens after fine-tuning and analyzing the model's performance under these conditions. This comparative analysis aims to reveal the extent to which embedding space robustness influences token-level resilience. The results of this token corruption assessment are detailed in \Cref{tab:corruption}. 
The most notable differences in robustness, particularly in resilience to token corruption, \textit{become more apparent at higher levels of token corruption}, specifically at $15$\% and $20$\%.

\begin{table}
\centering
\small
\begin{tabular}{llrrr}
\toprule
& & \multicolumn{3}{c}{Token corruption [\%]} \\
\cmidrule{3-5}
& & $10$ & $15$ & $20$ \\
\midrule
\multirow{4}{*}{\rotatebox[origin=c]{0}{BERT}}
& Base & $.721$ & $.703$ & $.671$   \\
& Jacobian & $.729$ & $.710$ & $.680$   \\
& H\"{o}lder & $.734$ & $.717$ & $.684$  \\
& \cellcolor{lightgray} \textsc{JacHess} & $\mathbf{.751}$ & $\mathbf{.742}$ & $\mathbf{.729}$   \\
\midrule
\multirow{4}{*}{\rotatebox[origin=c]{0}{OPT-125m}}
& Base & $.681$ & $.654$ & $.613$   \\
& Jacobian & $.694$ & $.667$ & $.642$   \\
& H\"{o}lder & $.693$ & $.679$ & $.648$  \\
& \cellcolor{lightgray} \textsc{JacHess} & $\mathbf{.731}$ & $\mathbf{.723}$ & $\mathbf{.704}$   \\
\midrule
\multirow{4}{*}{\rotatebox[origin=c]{0}{OPT-1.3b}}
& Base & $.814$ & $.802$ & $.743$ \\
& Jacobian & $.820$ & $.807$ & $.759$   \\
& H\"{o}lder & $.823$ & $.814$ & $.770$ \\
& \cellcolor{lightgray} \textsc{JacHess} & $\mathbf{.840}$ & $\mathbf{.831}$ & $\mathbf{.802}$ \\
\midrule
\multirow{4}{*}{\rotatebox[origin=c]{0}{OPT-6.7b*}}
& Base & $.831$ & $.809$ & $.787$   \\
& Jacobian & $.834$ & $.819$ & $.799$   \\
& H\"{o}lder & $.834$ & $.817$ & $.796$  \\
& \cellcolor{lightgray} \textsc{JacHess} & $\mathbf{.851}$ & $\mathbf{.839}$ & $\mathbf{.813}$   \\
\midrule
\multirow{4}{*}{\rotatebox[origin=c]{0}{LLaMA-2-7b*}}
& Base & $.831$ & $.813$ & $.786$   \\
& Jacobian & $.837$ & $.822$ & $.804$   \\
& H\"{o}lder & $.832$ & $.811$ & $.796$  \\
& \cellcolor{lightgray} \textsc{JacHess} & $\mathbf{.869}$ & $\mathbf{.853}$ & $\mathbf{.829}$   \\
\bottomrule
\end{tabular}
\caption{Average predictive accuracy with token corruption. We adjust the percentage of token corruption to $10$\%, $15$\%, and $20$\%. For each dataset, we conduct experiments five times using different seeds and report the average score on the GLUE benchmark. Best scores within the same model and token corruption setup are shown in \textbf{bold}.}
\label{tab:corruption}
\end{table}

\subsection{Generalization}

\begin{table*}
\centering
\small
\begin{tabular}{clrrrrrrrrr}
\toprule
& & CoLA & SST-2 & MRPC & STS-B & QQP & MNLI & QNLI & RTE & avg. \\
\midrule
\multirow{7}{*}{\rotatebox[origin=c]{90}{BERT}}
& Base & $.466$ & $.894$ & $.852$ & $.855$ & $.810$ & $.700$ & $.832$ & $.610$ & $.752$ \\
& Jacobian$_\text{train}$ & $.471$ & $.883$ & $.850$ & $.847$ & $.806$ & $.704$ & $.821$ & $.607$ & $.749$ \\
& Jacobian$_\text{val}$  & $.475$ & $.892$ & $.854$ & $.845$ & $.812$ & $.710$ & $.823$ & $.613$ & $.753$ \\
& Cross-H\"{o}lder$_\text{train}$ & $.498$ & $.901$ & $.842$ & $.839$ & $.824$ & $.707$ & $.830$ & $.605$ & $.756$ \\
& Cross-H\"{o}lder$_\text{val}$ & $.504$ & $.905$ & $.852$ & $.836$ & $\mathbf{.829}$ & $.712$ & $.838$ & $.616$ & $.762$ \\
& \cellcolor{lightgray} \textsc{JacHess}$_\text{train}$ & $.514$ & $\mathbf{.912}$ & $.848$ & $.862$ & $.816$ & $.710$ & $.836$ & $.621$ & $.765$ \\
& \cellcolor{lightgray} \textsc{JacHess}$_\text{val}$ & $\mathbf{.557}$ & $.906$ & $\mathbf{.864}$ & $\mathbf{.891}$ & $.828$ & $\mathbf{.723}$ & $\mathbf{.854}$ & $\mathbf{.643}$ & $\mathbf{.783}$ \\
\midrule
\multirow{7}{*}{\rotatebox[origin=c]{90}{OPT-125m}}
& Base & $.452$ & $.883$ & $.760$ & $.824$ & $.693$ & $.614$ & $.742$ & $.582$ & $.694$ \\
& Jacobian$_\text{train}$ & $.450$ & $.872$ & $.779$ & $.813$ & $.704$ & $.629$ & $.738$ & $.590$ & $.697$ \\
& Jacobian$_\text{val}$ & $.454$ & $.869$ & $.784$ & $.818$ & $.709$ & $.639$ & $.747$ & $.608$ & $.704$ \\
& Cross-H\"{o}lder$_\text{train}$ & $.461$ & $.881$ & $.758$ & $.830$ & $.722$ & $.661$ & $.744$ & $.614$ & $.709$  \\
& Cross-H\"{o}lder$_\text{val}$ & $.470$ & $.880$ & $.771$ & $.839$ & $.731$ & $.657$ & $.751$ & $.620$ & $.715$  \\
& \cellcolor{lightgray} \textsc{JacHess}$_\text{train}$ & $\mathbf{.474}$ & $\mathbf{.896}$ & $.819$ & $.825$ & $.736$ & $.672$ & $.757$ & $.610$ & $.724$ \\
& \cellcolor{lightgray} \textsc{JacHess}$_\text{val}$  & $.470$ & $.884$ & $\mathbf{.835}$ & $\mathbf{.848}$ & $\mathbf{.768}$ & $\mathbf{.691}$ & $\mathbf{.787}$ & $\mathbf{.628}$ & $\mathbf{.739}$ \\
\midrule
\multirow{7}{*}{\rotatebox[origin=c]{90}{OPT-1.3b}}
& Base & $.601$ & $.945$ & $.905$ & $.913$ & $.847$ & $.755$ & $.903$ & $.742$ & $.826$ \\
& Jacobian$_\text{train}$ & $.589$ & $.940$ & $.911$ & $.908$ & $.851$ & $.770$ & $.892$ & $.731$ & $.824$ \\
& Jacobian$_\text{val}$ & $.598$ & $.943$ & $.909$ & $\mathbf{.916}$ & $.857$ & $.779$ & $.894$ & $.742$ & $.830$ \\
& Cross-H\"{o}lder$_\text{train}$ & $.612$ & $.939$ & $.910$ & $.902$ & $.879$ & $.773$ & $\mathbf{.924}$ & $.747$ & $.836$  \\
& Cross-H\"{o}lder$_\text{val}$ & $.608$ & $.949$ & $.909$ & $.907$ & $.884$ & $.771$ & $.921$ & $.750$ & $.837$ \\
& \cellcolor{lightgray} \textsc{JacHess}$_\text{train}$ & $.610$ & $.948$ & $.913$ & $.903$ & $.863$ & $.803$ & $.918$ & $.745$
& $.838$ \\
& \cellcolor{lightgray} \textsc{JacHess}$_\text{val}$ & $\mathbf{.614}$ & $\mathbf{.955}$ & $\mathbf{.919}$ & $.908$ & $\mathbf{.892}$ & $\mathbf{.811}$ & $.921$ & $\mathbf{.751}$ & $\mathbf{.846}$ \\
\midrule
\multirow{7}{*}{\rotatebox[origin=c]{90}{OPT-6.7b*}}
& Base & $.652$ & $.951$ & $.908$ & $.916$ & $.869$ & $.797$ & $.907$ & $.750$ & $.844$ \\
& Jacobian$_\text{train}$ & $.649$ & $.940$ & $.912$ & $.909$ & $.873$ & $.822$ & $.913$ & $.747$ & $.846$ \\
& Jacobian$_\text{val}$ & $.652$ & $.944$ & $.914$ & $.907$ & $.879$ & $.831$ & $.917$ & $.752$ & $.850$ \\
& Cross-H\"{o}lder$_\text{train}$ & $.654$ & $.949$ & $.914$ & $.903$ & $.881$ & $.814$ & $.901$ & $.741$ & $.845$  \\
& Cross-H\"{o}lder$_\text{val}$ & $.650$ & $.959$ & $.920$ & $.907$ & $.887$ & $.821$ & $.907$ & $.745$ & $.850$  \\
& \cellcolor{lightgray} \textsc{JacHess}$_\text{train}$ & $.651$ & $.953$ & $.919$ & $.911$ & $.889$ & $.827$ & $.918$ & $.750$ & $.852$ \\
& \cellcolor{lightgray} \textsc{JacHess}$_\text{val}$ & $\mathbf{.688}$ & $\mathbf{.959}$ & $\mathbf{.928}$ & $\mathbf{.922}$ & $\mathbf{.907}$ & $\mathbf{.852}$ & $\mathbf{.929}$ & $\mathbf{.776}$ & $\mathbf{.870}$ \\
\midrule
\multirow{7}{*}{\rotatebox[origin=c]{90}{LLaMA-2-7b*}}
& Base & $.691$ & $.957$ & $.912$ & $.924$ & $.910$ & $.843$ & $.925$ & $.781$ & $.868$ \\
& Jacobian$_\text{train}$ & $.681$ & $.940$ & $.893$ & $.903$ & $.882$ & $.837$ & $.912$ & $.764$ & $.852$ \\
& Jacobian$_\text{val}$ & $.693$ & $.955$ & $.915$ & $.913$ & $.890$ & $.844$ & $.923$ & $.769$ & $.863$ \\
& Cross-H\"{o}lder$_\text{train}$ & $.688$ & $.951$ & $.909$ & $.915$ & $.914$ & $.832$ & $.927$ & $.759$ & $.862$  \\
& Cross-H\"{o}lder$_\text{val}$ & $.691$ & $.949$ & $.913$ & $.917$ & $.909$ & $.838$ & $.931$ & $.779$ & $.866$ \\
& \cellcolor{lightgray} \textsc{JacHess}$_\text{train}$ & $.712$ & $.962$ & $.908$ & $.921$ & $.919$ & $.851$ & $.933$ & $.798$ & $.876$ \\
& \cellcolor{lightgray} \textsc{JacHess}$_\text{val}$ & $\mathbf{.746}$ & $\mathbf{.973}$ & $\mathbf{.951}$ & $\mathbf{.934}$ & $\mathbf{.929}$ & $\mathbf{.872}$ & $\mathbf{.940}$ & $\mathbf{.813}$ & $\mathbf{.895}$ \\
\bottomrule
\end{tabular}
\caption{Predictive accuracy on the GLUE benchmark. We use Matthew's correlation for CoLA, $F_1$ for MRPC and QQP, Spearman's correlation for STS-B, and accuracy for the rest. We report the average performance with five different seeds for each dataset.  In the last column, we report the average across the datasets for each model. The highest scores for each dataset are shown in \textbf{bold}.}
\label{tab:robgen}
\end{table*}

Following the evaluation of robustness, we shift our focus to evaluating the impact of our methodologies on model generalization. This evaluation, acting as a benchmark for standard generalization performance, sets the stage for understanding how robustness enhancements in embedding and token-level spaces contribute to the overall generalization capabilities of the models.

Our findings, detailed in \Cref{tab:robgen}, indicate that \textit{\jachess{} consistently outperforms both standard unregularized fine-tuning and other regularization baselines}, affirming the efficacy of our approach in bolstering model generalization through enhanced robustness. Moreover, \jachess{} provides consistent improvements as the scale of the models increases. The relative improvement is notably pronounced in the larger models, namely OPT-6.7b and LLaMA-2-7b.

\section{Uncertainty Calibration}


Building on existing research suggesting that smoothness in model representations concerning inputs can lead to more dependable uncertainty quantification \cite{rosca-etal-2020-case}, we investigate the efficacy of our method in this critical area. To this end, we employ the Brier score as our metric, chosen for its comprehensive nature. The Brier score encompasses aspects of calibration and refinement in the evaluation of uncertainty quantification, measuring the mean squared distance between predicted probabilities and actual outcomes. Functioning as a loss function, a lower Brier score indicates better performance.

\begin{figure*}
\centering
\begin{subfigure}{.32\linewidth}
  \centering
  \includegraphics[width=\linewidth]{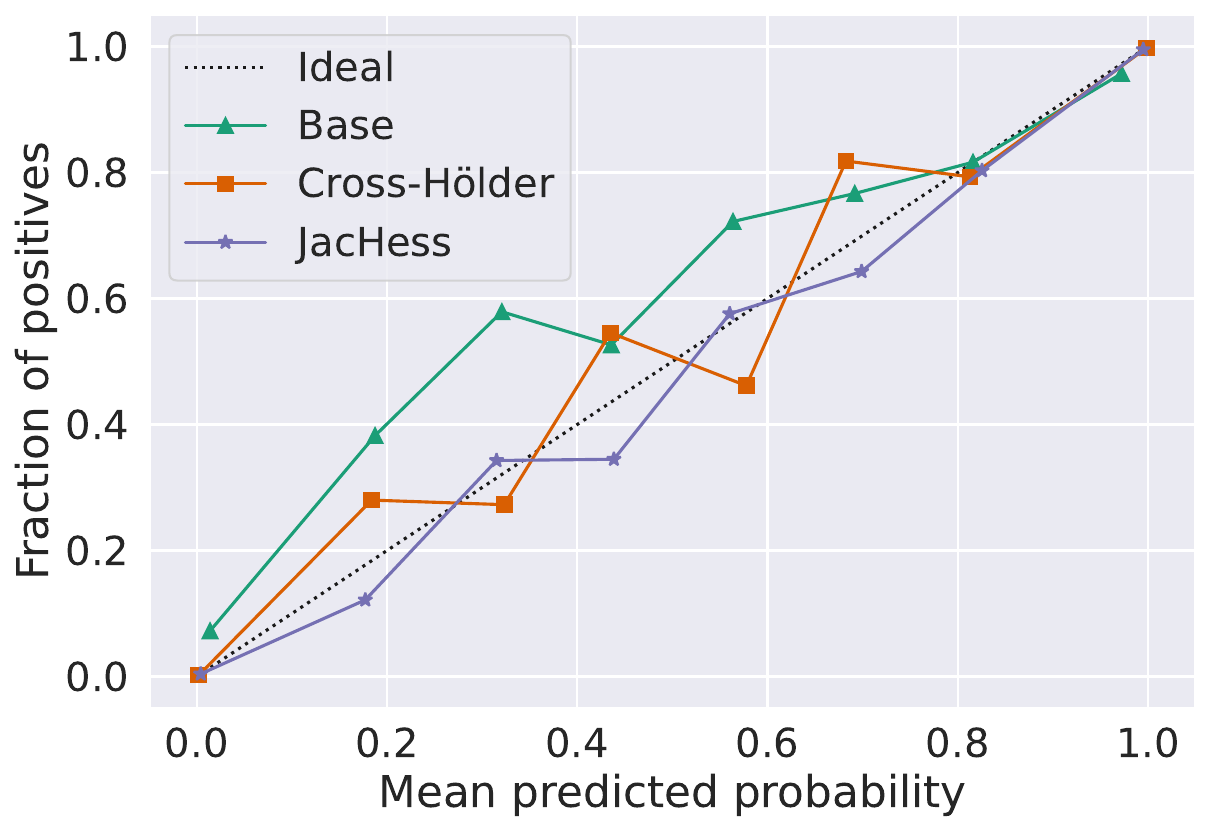}
  \caption{CoLA}
  \label{fig:cal-cola}
\end{subfigure}
\begin{subfigure}{.32\linewidth}
  \centering
  \includegraphics[width=\linewidth]{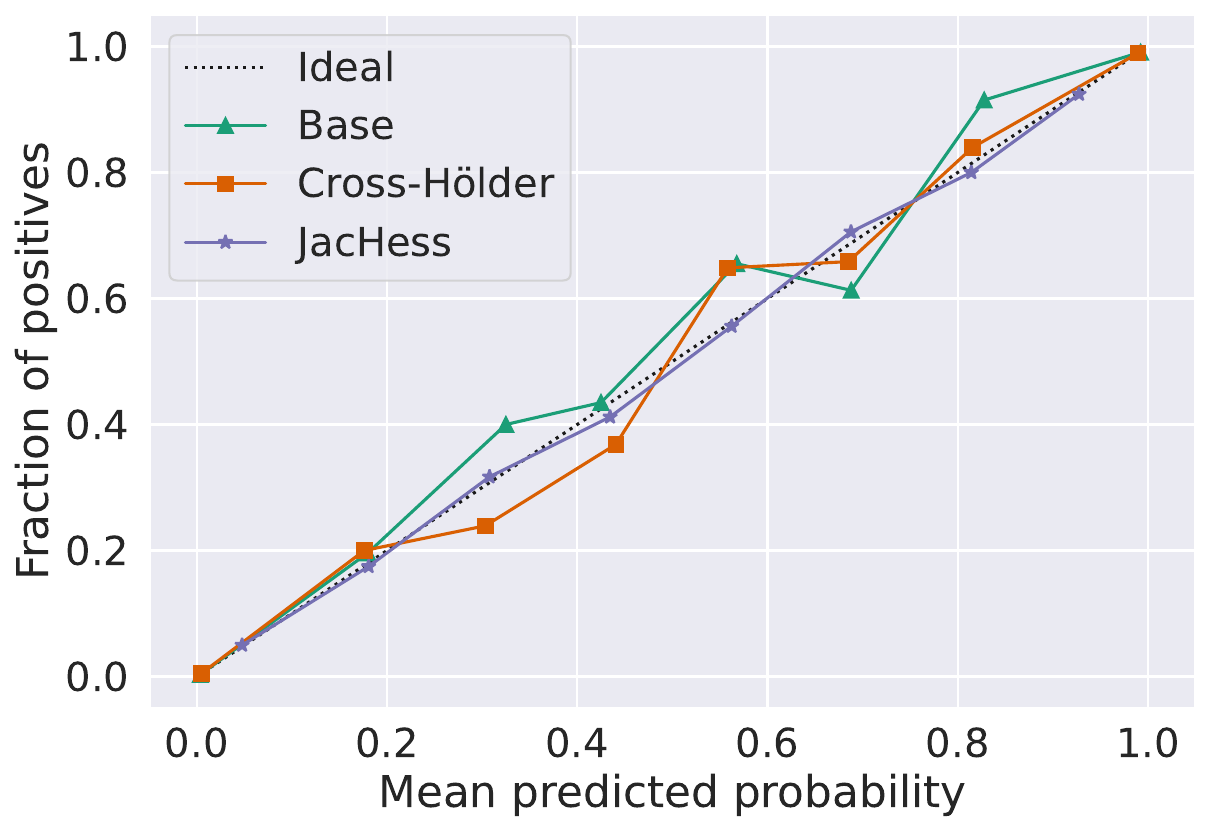}
  \caption{SST-2}
  \label{fig:cal-sst}
\end{subfigure}
\begin{subfigure}{.32\linewidth}
  \centering
  \includegraphics[width=\linewidth]{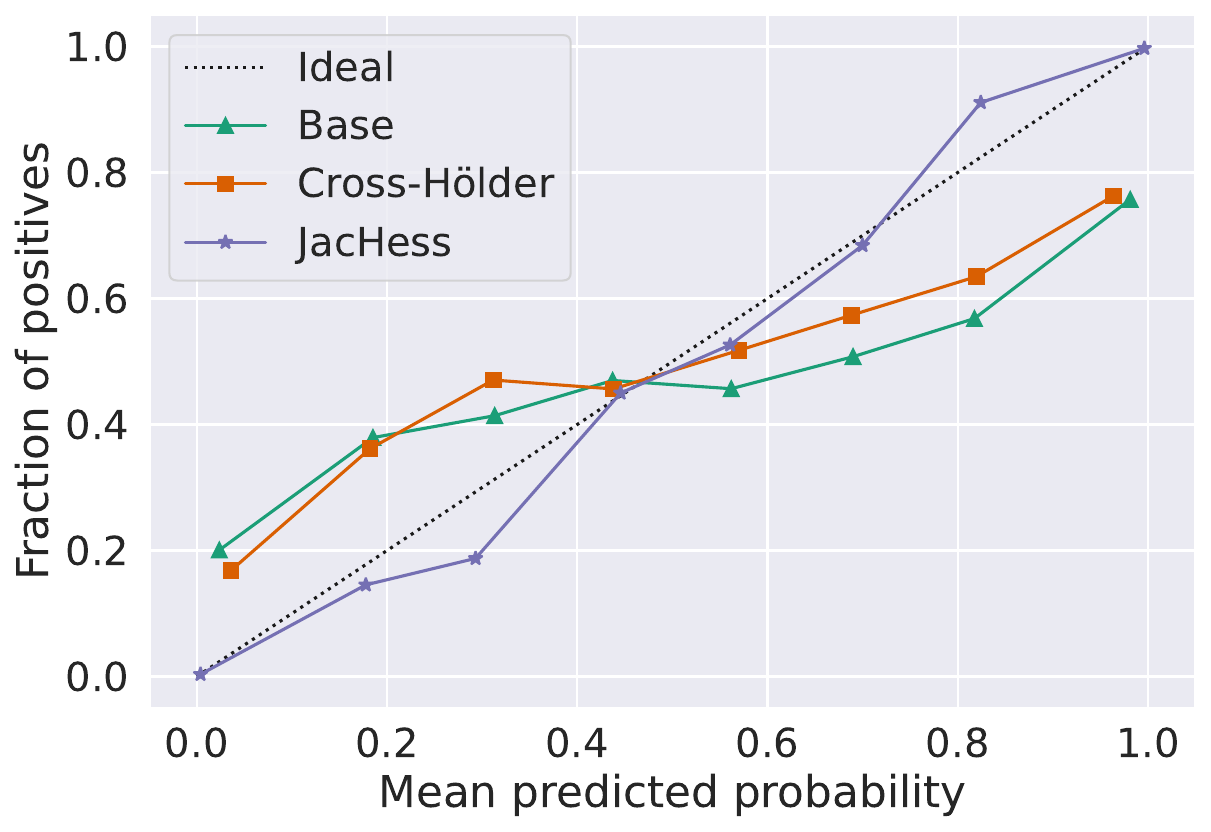}
  \caption{MRPC}
  \label{fig:cal-mrpc}
\end{subfigure}
\begin{subfigure}{.32\linewidth}
  \centering
  \includegraphics[width=\linewidth]{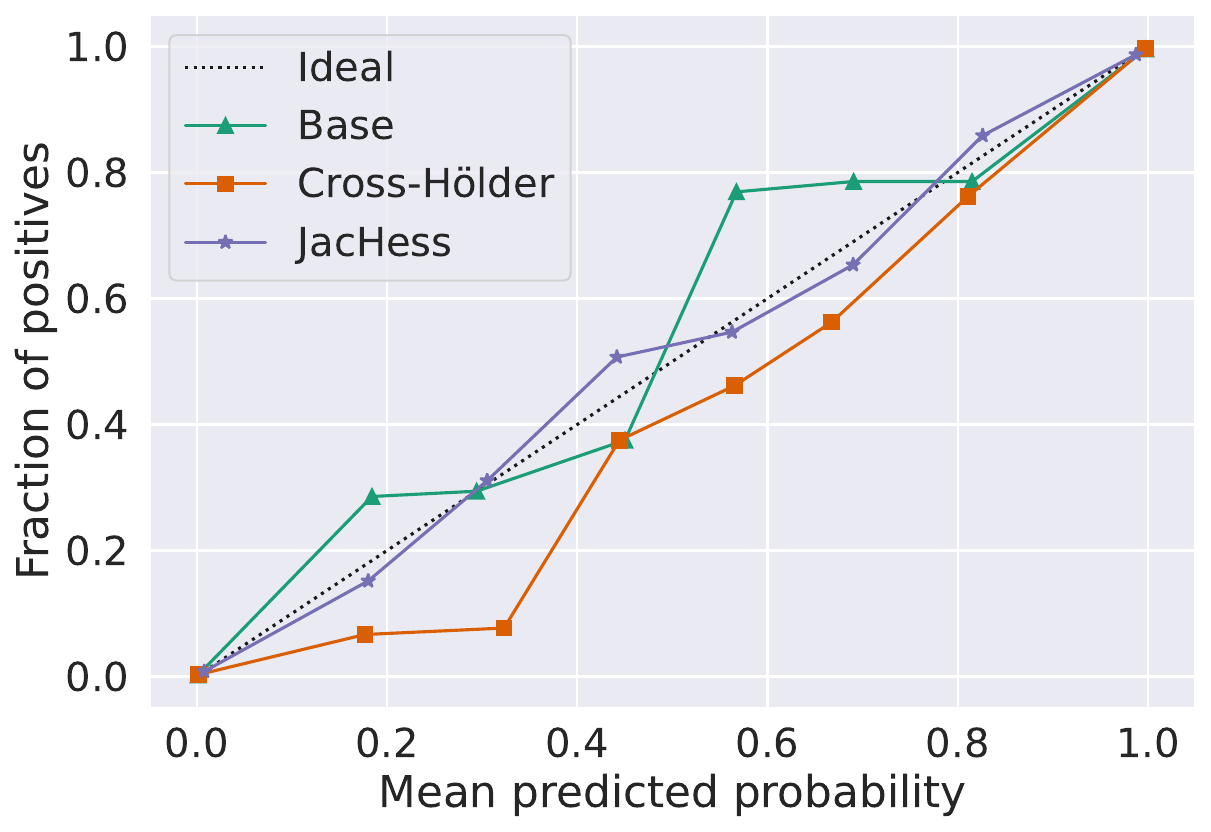}
  \caption{RTE}
  \label{fig:cal-rte}
\end{subfigure}
\begin{subfigure}{.32\linewidth}
  \centering
  \includegraphics[width=
  \linewidth]{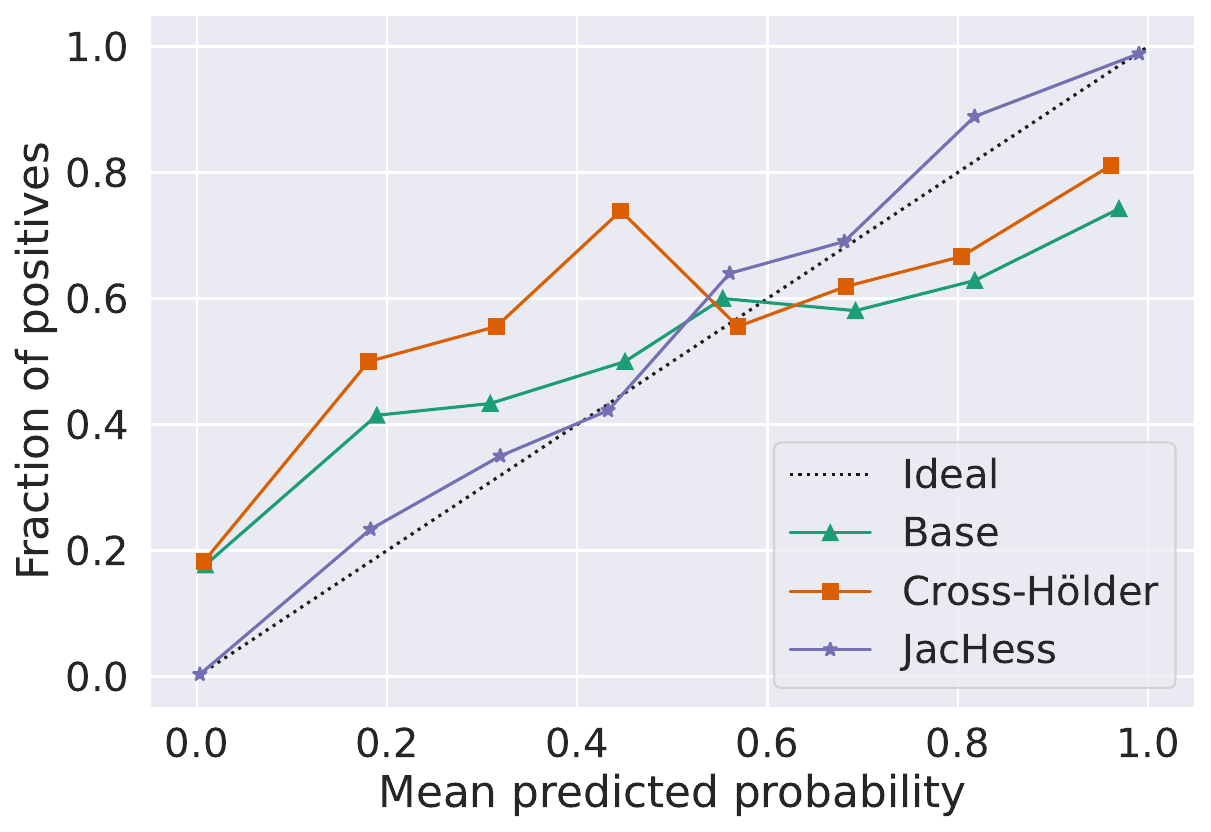}
  \caption{QQP}
  \label{fig:cal-qqp}
\end{subfigure}
\begin{subfigure}{.32\linewidth}
  \centering
  \includegraphics[width=
  \linewidth]{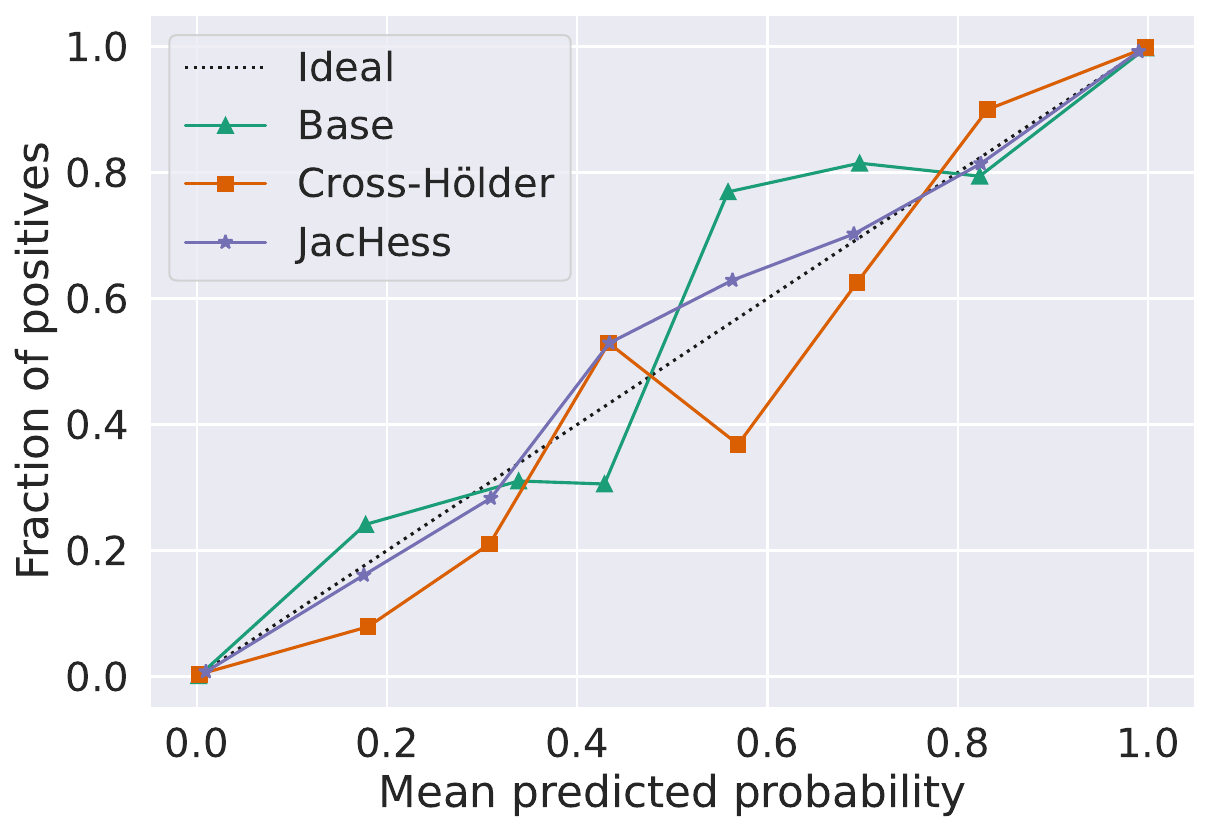}
  \caption{QNLI}
  \label{fig:cal-qqp}
\end{subfigure}
\caption{Calibration plots for binary classification datasets. We accumulate the results in all five different seeds and use eight bins for mean predicted probability. We show the calibration plots for LLaMA-2-7b.}
\label{fig:cal}
\end{figure*}

We report the Brier scores for different models in \Cref{tab:brier}. Besides acting as a generalization enhancer, \jachess{} also contributes to better uncertainty estimation, which results in lower Brier loss, while \Cref{fig:cal} presents the calibration scores, offering insights into our method's performance in ensuring reliable uncertainty quantification in predictive modeling. The findings indicate that \textit{\jachess{} surpasses baseline methods, offering superior Brier scores and enhanced calibration, underscoring its efficacy in rendering reliable uncertainty estimates in models}.

\begin{table}
\centering
\small
\begin{tabular}{l|ccccc}
\toprule
Model & Base & Cross-H\"{o}lder & \cellcolor{lightgray} \textsc{JacHess} \\
\midrule
BERT & $.213$ & $.202$ & $\mathbf{.184}$ \\
OPT-125m & $.256$ & $.233$ & $\mathbf{.204}$ \\
OPT-1.3b & $.184$ & $.193$ & $\mathbf{.157}$ \\
OPT-6.7b* & $.189$ & $.157$ & $\mathbf{.094}$ \\
LLaMA-2-7b* & $.167$ & $.163$ & $\mathbf{.089}$ \\
\bottomrule
\end{tabular}
\caption{Average Brier scores across binary classification tasks. We report the average Brier scores for the CoLA, SST-2, MRPC, RTE, QQP, and QNLI datasets, all of which involve binary classification tasks. Notably, lower Brier scores indicate better performance. Best scores are indicated in \textbf{bold}.}
\label{tab:brier}
\end{table}

\section{Analysis}
\label{sec:analysis}

\begin{table*}
\centering
\begin{tabular}{l|ccccc}
\toprule
Strategy & BERT & OPT-125m & OPT-1.3b & OPT-6.7b* & LLaMA-2-7b* \\
\midrule
Penultimate layer (logits) & $.743$ & $.709$ & $.832$ & $.851$ & $.874$ \\
Uniform throughout layers & $.775$ & $.726$ & $.845$ & $.868$ & $.883$ \\
Inverse to base smoothness & $.733$ & $.692$ & $.803$ & $.811$ & $.819$ \\
Normalized base smoothness & $.778$ & $.735$ & $\mathbf{.848}$ & $.865$ & $.892$ \\
Softmax base smoothness & $\mathbf{.783}$ & $\mathbf{.739}$ & $.846$ & $\mathbf{.870}$ & $\mathbf{.895}$ \\
\bottomrule
\end{tabular}
\caption{Comparison of strategies for regularization at different application points evaluated on the GLUE benchmark. We report the average performance across datasets for each model, where we run each experiment five times with different seeds. \textit{Penultimate layer (logits)} involves regularizing only the logits Hessians' norms. \textit{Uniform throughout layers} means applying regularization uniformly across all layers. \textit{Inverse to base smoothness} sets $\boldsymbol{\lambda}$ inversely proportional to the base model's smoothness, and \textit{Normalized base smoothness} aligns $\boldsymbol{\lambda}$ directly with the base model's smoothness, while \textit{softmax base smoothness} applies softmax instead of normalization. Here, smoothness is measured as the inverse of the Jacobians' norms.}
\label{tab:ablation}
\end{table*}

\begin{table*}
\centering
\begin{tabular}{c|ccccc}
\toprule
\# of estimated Hessian norms  & BERT & OPT-125m & OPT-1.3b & OPT-6.7b* & LLaMA-2-7b* \\
\midrule
0 & $.764$ & $.698$ & $.821$ & $.849$ & $.873$ \\
5 & $.771$ & $.721$ & $.839$ & $.854$ & $.886$ \\
10 & $\mathbf{.783}$ & $\mathbf{.739}$ & $.846$ & $.868$ & $\mathbf{.895}$ \\
20 & $.781$ & $.737$ & $\mathbf{.850}$ & $\mathbf{.870}$ & $.892$ \\
50 & $.754$ & $.713$ & $.831$ & $.857$ & $.880$ \\
\bottomrule
\end{tabular}
\caption{Comparison of \jachess{} flavors based on the number of Hessian norms. The number of norms corresponds to the number of sampled dimensions for a particular layer output for which the Hessian norm is estimated.}
\label{tab:ablation}
\end{table*}

In this section, we aim to validate the design choices behind \jachess{}.

\subsection{Span and degree of regularization}
Our initial investigation focuses on understanding how the \textbf{span} of regularization impacts performance. We compare the effects of applying our regularization method to the logits versus its application across multiple network layers. Additionally, we evaluate the impact of the \textit{degree} of regularization on different layers by controlling the regularization coefficients $\boldsymbol{\lambda}$. As shown in \Cref{tab:ablation}, using a base smoothness distribution (inversely related to Jacobian norms) for initializing the regularization factors emerges as the most effective strategy. Applying uniform regularization across all layers surpasses the strategy of only regularizing the logits. In contrast, initializing the regularization coefficients inversely proportionate to the base smoothness for each layer detrimentally affects performance.

\subsection{How many estimated Hessian norms?}
Continuing our analysis, we examine how the regularization of varying numbers of layer representation dimensions affects model performance. For each specific dimension, we calculate the Hessian in relation to the inputs, a process that is notably computation-intensive. Therefore, we opt for the random selection of a subset of dimensions for each batch. We report the results in \Cref{tab:ablation}. Our observations indicate a positive impact on performance when more than $5$ dimensions are selected, with $10$ dimensions being the default choice in our approach. However, expanding this selection to $20$ dimensions leads to a performance plateau, and extending it further to $50$ dimensions results in performance degradation, which is due to excessive smoothing of the layer outputs.

\section{Conclusion}
The importance of robustness and generalization in pre-trained language models (PLMs) cannot be overstated, as these qualities significantly influence model reliability and efficacy. Our research offers a comprehensive empirical analysis highlighting how input smoothness enforcement via Jacobian- and Hessian-based regularization within the embedding space enhances these aspects. We introduce \jachess{}, an innovative regularization framework that not only surpasses standard fine-tuning and existing methods in improving model generalization but also bolsters the models' ability to quantify uncertainty, thereby yielding more reliable predictions. This study not only sheds light on the critical role of representation smoothness in model robustness and generalization but also presents \jachess{} as an approach that enriches both the theoretical and practical landscape of PLM optimization.


\bibliography{tacl2021}
\bibliographystyle{acl_natbib}

\end{document}